%% file: camera_ready.tex
\documentclass{article}

    \PassOptionsToPackage{numbers, compress}{natbib}
 \usepackage[position, final]{neurips_2026}
\usepackage[utf8]{inputenc} % allow utf-8 input
\usepackage[T1]{fontenc}    % use 8-bit T1 fonts
\usepackage[colorlinks=true, urlcolor=utburnt, citecolor=utburnt, linkcolor=utburnt]{hyperref}       % hyperlinks
\usepackage{url}            % simple URL typesetting
\usepackage{booktabs}       % professional-quality tables
\usepackage{amsfonts}       % blackboard math symbols
\usepackage{nicefrac}       % compact symbols for 1/2, etc.
\usepackage{microtype}      % microtypography
\usepackage{xcolor}         % colors

\usepackage{algorithm}
\usepackage{kotex}
\usepackage{multirow}
\usepackage{makecell} 
\usepackage{graphicx} 
\usepackage{adjustbox}
\usepackage{float}
\usepackage{wrapfig}
\usepackage{colortbl}
\usepackage[endLComment=,italicComments=false]{algpseudocodex}
\usepackage{algorithm}
\input{insbox.tex}

\usepackage{amsmath}
\usepackage{amssymb}
\usepackage{mathtools}
\usepackage{amsthm}

\usepackage{xcolor}

\usepackage{pifont}% http://ctan.org/pkg/pifont
\usepackage{xspace}
\usepackage{dsfont}

\usepackage{amsmath,amsfonts}

\definecolor{myred}{HTML}{F54254}
\definecolor{myorange}{HTML}{FFB135}
\definecolor{mygreen}{HTML}{10BD35}
\definecolor{myblue}{HTML}{598BE7}
\definecolor{mypurple}{HTML}{9A1C6B}
\definecolor{plgray}{HTML}{999999}
\definecolor{ut}{HTML}{CC5500}
\definecolor{utburnt}{HTML}{BF5700}

\newcommand{\mathbbm}[1]{\text{\usefont{U}{bbm}{m}{n}#1}}

\usepackage{lipsum}
\usepackage{comment}
\theoremstyle{plain}
\newtheorem{theorem}{Theorem}[section]
\newtheorem{proposition}[theorem]{Proposition}

\theoremstyle{definition}

\theoremstyle{remark}

\definecolor{lightcyan}{rgb}{0.88, 1, 1}
\definecolor{asparagus}{rgb}{0.53, 0.66, 0.42}
\definecolor{azure}{rgb}{0.0, 0.5, 1.0}
\definecolor{brightpink}{rgb}{1.0, 0.0, 0.5}
\definecolor{boston}{rgb}{0.8, 0.0, 0.0}
\definecolor{gray}{rgb}{0.75, 0.75, 0.75}
\definecolor{lightgray}{rgb}{0.88, 0.88, 0.88}
\definecolor{darkgray}{rgb}{0.50, 0.50, 0.50}
\definecolor{orange2}{rgb}{0.99, 0.86, 0.70}
\definecolor{pastelgray}{rgb}{0.81, 0.81, 0.77}
\definecolor{orangered}{rgb}{.99, .40, .00}
\definecolor{darkUT}{HTML}{BF5700}
\definecolor{lightUT}{HTML}{FFF1E6}
\definecolor{darkUT_B}{HTML}{005F86}
\definecolor{lightUT_B}{HTML}{F5FDFF}
\definecolor{darkUT_G}{HTML}{333F48}
\definecolor{lightUT_G}{HTML}{F5F4F0}
\definecolor{darkgreen}{HTML}{F5F4F0}
\definecolor{apricot}{HTML}{FFD7B8}

\usepackage[most]{tcolorbox}

\usepackage{tikz}

\newcounter{takeawayonly}

\newcommand{\parag}[1]{\vspace{+0.0mm}\noindent\textbf{#1}}
\newcommand{\takeawayonly}[1]{
    \vspace{-0.05cm}
    \refstepcounter{takeawayonly}
    \begin{tcolorbox}[
        colback=darkUT!8,                       
        colframe=darkUT!95,                     
        arc=4pt,                    % Less rounded corners for a modern look
        boxsep=5pt,                 % Inner margin between text and frame
        left=2pt,                  % Left margin within the box
        right=2pt,                 % Right margin within the box
        top=4pt,                    % Top margin within the box
        bottom=4pt,                 % Bottom margin within the box
        boxrule=0.8pt,              % Frame thickness
        drop shadow=gray!30!white,  % Subtle shadow effect for depth
        enhanced jigsaw             % Allows for the shadow effect and better arcs
    ]
    \vspace{-0.15cm}
        \parag{\textbf{\textit{Core Claim:}}} #1
    \vspace{-0.15cm}
    \end{tcolorbox}
}

\title{Position: Beyond Training Time, Test-Time Coordination is Essential for Cooperative MARL}

\author{%
 Dongsu Lee\thanks{Equal contribution.}\:\:$^{1}$ \quad 
 Sooraj Sathish\footnotemark[1]\:\:$^{1}$ \quad
 Joonkyung Kim$^{2}$ \\ 
 \textbf{Woojun Kim}$^3$ \quad
 \textbf{Amy Zhang}$^1$ 
 \\
 $^{1}$The University of Texas at Austin \quad $^{2}$Texas A\&M University \quad $^{3}$KAIST
 \\
 \texttt{\{dongsu.lee, ss242758\}@utexas.edu} \\
}

\begin{document}

\maketitle

\begin{abstract}
We argue that cooperative multi-agent reinforcement learning (MARL) should treat test-time coordination as part of the problem definition rather than a deployment artifact. Our focus is on long-horizon, centrally monitored cooperative systems such as warehouse robotics, automated storage-and-retrieval systems, and cloud-resource scheduling, where teams share a reward, operate under shifting objectives, and already run with global telemetry at deployment. The dominant paradigm in this regime, centralized training with decentralized execution (CTDE), uses global information only during training and provides no coordination channel at execution. This is sufficient when agents are weakly coupled or the environment is stationary, but it fails when the team must realign after conditions change. The limitation is structural rather than algorithmic, and more training cannot add a channel that does not exist at execution time. We call this the \textbf{recoverability gap}. To address it, we name the underexplored design space of intermittent test-time coordination as \textbf{centralized training with mixed execution (CTME)}, in which a coordinator observes global state and emits high-level signals while agents act locally. CTME is not a particular learning algorithm, but a cooperative MARL problem setting that makes test-time information structure explicit. It specifies team-level information, high-level signals, and update frequency after training, with CTDE recovered as a limit case. We do not argue that CTDE is obsolete. We argue that removing the test-time channel is a modeling choice, not a necessity, and the wrong default when global telemetry is available and teams must recover from drift.
\end{abstract}

\setcounter{tocdepth}{1}
\addtocontents{toc}{\protect\setcounter{tocdepth}{-1}}

\section{Introduction}
\label{sec:intro}
\textbf{What is our scope?} Cooperative MARL has matured rapidly, but largely on simulated benchmarks~\cite{samvelyan2019starcraftmultiagentchallenge, papoudakis2021benchmarking}. The settings that genuinely call for cooperation are more specific: teams run by a single operator that must keep working together as conditions change, \textit{e.g.}, warehouse robotics~\cite{krnjaic2024scalable}, traffic networks~\cite{wei2019colight, chu2019multi}, and cloud resource scheduling~\cite{nagarajan2025multi}, where coordination challenges arise from \textit{shifting objectives}, \textit{dynamic role reassignment}, and \textit{long horizons}. Our position lies in settings of this kind, in which agents share an environment and centralized training is available and natural~\cite{foerster2016learning, yu2022surprising}. We do not consider settings such as autonomous driving, where each agent plans for itself in a shared environment and the problem reduces to single-agent RL with zero-shot coordination~\cite{barrett2011empirical, wang2024zsc, lee2024episodic}. Thus, our primary question is that of how much of the training-time information should remain at execution. 

\textbf{Why is coordination a structural problem?} 
Coordination is the management of dependencies among agents~\cite{malone1994interdisciplinary}. Fully decoupled agents forming a team can each be trained on their own, possibly even asynchronously. Weakly coupled agents are a good approximation to this ideal \cite{simon2012architecture}. On the other hand, as coupling grows, the right action for one agent depends increasingly on what the others do and on team-level conditions that change over time. Such reciprocal interdependence calls for ongoing mutual adjustment rather than standing rules. Thus, under drift, coordination must arise from the information flow at execution time rather than from training alone.

\textbf{Comparison to fully centralized and decentralized execution}
\textit{Centralized execution} makes the joint action a single decision. This guarantees coordination in principle, but the joint action space grows exponentially with team size. In addition, although the central controller receives all local observations as input, it can overlook local details (\textit{which are not directly related to the global goal}), so it is difficult to control each agent at this level~\cite{brooks1986robust}. On the other hand, in \textit{decentralized execution}  each agent acts independently. This avoids the cost, but introduces new failure modes. Over long horizons, each agent operates with only a partial view of the team state, so locally reasonable decisions can compound into team-level misalignment~\cite{li2025revisiting, lee2026recipe}. The range of objectives a single trained team can support is also bounded by what each local policy can implicitly encode.

The dominant compromise is CTDE~\cite{oliehoek2008optimal}, which uses global information during training to address credit assignment (\textit{e.g.}, value decomposition~\cite{sunehag2017value, son2019qtran} or centralized critics~\cite{lowe2017multi}) but reverts to fully decentralized execution. CTDE has been remarkably successful as a training paradigm and is the right default in many settings, particularly when centralized signals are available during training but absent at deployment. However, in centrally monitored systems where global telemetry is available even during deployment, CTDE's restriction on agent execution becomes a limitation. This inherits the failure modes of decentralized execution, leaving the team with no channel through which to realign when objectives shift, bottlenecks emerge, or role assignments grow stale~\cite{hernandez2017survey, papoudakis2019dealing, nekoei2021continuous}. We call this absence of a test-time realignment mechanism the \textit{recoverability gap}.
 
\takeawayonly{\textit{In long-horizon cooperative deployments, coordination is not a one-time outcome of training but a property the team must actively maintain during execution. CTDE provides a training mechanism for coordination but no execution mechanism, leaving the team unable to realign once conditions evolve.}}

\textbf{How have prior works considered this problem?}  Communication-based MARL~\cite{foerster2016learning, niu2021multi, lee2025unifying} exists precisely because acting on local observations alone is inadequate in coupled teams, and the growth of this subfield indicates that many tasks need more than a no-channel default. However, communication faces a dilemma. If agents can reach the entire team at execution, the deployment already has the infrastructure to assemble team-level information at one node, and when communication is free and complete, the decentralized problem collapses into a centralized one~\cite{pynadath2002communicative, goldman2004decentralized}. If communication is only local, each agent sees a wider neighborhood, but the problem remains decentralized and information beyond it stays aliased. In centrally monitored systems, the team-level picture is already assembled by the operator's monitoring layer, so rebuilding it bottom-up from local messages is a choice the problem does not require. Thus, the real question is not \textit{whether} a test-time channel is needed, but \textit{which} channel, carrying \textit{what}, and \textit{how often}.
 
\textbf{We argue that closing the recoverability gap requires test-time coordination as a structural primitive of cooperative MARL.} When team-level telemetry exists at deployment, as in the inventory management of supply chain~\cite{eisenach2024neural}, discarding it is an inefficient choice rather than a constraint of the environment. We call the underexplored design space of intermittent test-time coordination \textit{centralized training with mixed execution} (CTME). In this setting, the coordinator observes the team-level state and intermittently emits high-level signals such as sub-goals, skills, roles, or latent intents; between activations, agents act on their local observations and the latest signal. CTME is a problem setting, not a learning algorithm. As a prior connection, CTDE, CTCE, and CTME share centralized training and differ only in what survives to execution: CTDE is the special case with no signal, and CTCE coincides with CTME only for primitive-action signals issued at every step.

Our position rests on three contributions. First, we identify the recoverability gap as a test-time information-structure problem that cannot be closed by training-time techniques alone. Second, we formalize CTME as a design space parameterized by signal abstraction and coordination frequency, containing CTDE as a special case while keeping CTCE structurally separate. Third, we use a Nash-versus-signal-conditioned policy gap for why test-time signals enable state-contingent recovery, not merely a static welfare improvement, and outline research directions for studying CTME.\footnote{Throughout, \textit{`test-time'} refers to the information that reaches the agents during execution, not to parameter updates like test-time adaptation. In Section~\ref{sec:ctme}, we clearly clarify \textit{what is `test-time'} for CTME.}

\section{Rationale for the Position}
\label{sec:rationale}
We support our position with three pillars: a domain analysis, a game-theoretic analysis, and a historical account of why this regime has been overlooked.

\subsection{When CTDE is the wrong default?}
\label{subsec:domain-analysis}
The argument for test-time coordination is sharpest in settings where centralized infrastructure is already part of the deployment. The operator possesses global state, the team's objective is shared, and operational systems monitor the team during execution. We restrict our analysis to such centrally trainable cooperative domains, which include the examples cited in Section~\ref{sec:intro}. What makes the CTDE default particularly inefficient here is that even paradigms originally motivated by inter-agent coordination, such as communication-based MARL, are trained jointly with local policies and leave the centralized signaling channel that operational systems already provide unused.
 
The level of coordinator authority should depend on the domain and maps onto the equilibrium concepts in Section~\ref{subsec:game-theory}. When agent interactions are weak or communication is unreliable, standard CTDE is appropriate, and the relevant outcome is a Nash equilibrium (NE)~\cite{holt2004nash}. In volatile and safety-critical domains, the coordinator provides guidance while agents retain override authority (\textit{e.g.}, congestion alerts in traffic networks), so each agent decides after seeing its recommendation, as in a correlated equilibrium (CE)~\cite{aumann1974subjectivity, aumann1987correlated}. In tightly coupled environments, authority is closer to a binding assignment (\textit{e.g.}, a warehouse rerouting an agent along a locally suboptimal route to avoid congestion), so agents commit before seeing the recommendation, as in a coarse correlated equilibrium (CCE)~\cite{moulin1978strategically}.
 
\subsection{From a broader equilibrium set to test-time recoverability}
\label{subsec:game-theory}
In cooperative settings, team decision theory holds that the decisive question is who knows what, and when~\cite{marschak1974decision}. Under decentralized execution, each agent acts on its local observation alone, and we use NE~\cite{nash2024non, holt2004nash} to model the resulting stable outcomes; the limitation is one of information structure, not of the learning algorithm. A shared signal enlarges the set of achievable outcomes, as formalized by the correlated equilibrium~\cite{aumann1974subjectivity, forges1986approach, aumann1987correlated}: a mediator, which is not a player and never acts on the environment, draws a joint action from some distribution and privately recommends each component to the corresponding agent. If no agent benefits from ignoring its recommendation, the distribution is a CE. Every NE induces a CE, but the set of Correlated equilibria is often strictly larger~\cite{aumann1974subjectivity, aumann1987correlated}. The weaker CCE~\cite{moulin1978strategically, young2004strategic} requires only that committing to the mediator be beneficial on average, and its set contains the CEs. Both are also easier to compute than NE~\cite{daskalakis2009complexity, papadimitriou2008computing}, but the key point for us is informational: shared signals expand the set of implementable joint behaviors.

We use CE and CCE as a structural lens rather than as what CTME computes. In CE, the mediator recommends primitive actions, and its incentive constraints concern strategic deviation. Neither holds for CTME: its coordinator sends a coordination signal, and all agents share one objective. What carries over is that a shared external signal enables joint behaviors that no profile of independent best responses can reach. The CE model is static: one mediator uses one fixed signaling distribution for one game. In long-horizon, non-stationary settings, the same idea has a stronger consequence. At test time, the coordinator observes the current team-level state and selects which coordinated outcome the team realizes. That is, the signal changes with the state. When objectives shift, or bottlenecks emerge, the team is moved from one outcome to another, a transition that decentralized agents cannot reliably make from local observations. This is what makes state-contingent recovery possible.
 
\subsection{Why this regime remained underexplored?}
\label{subsec:why-underexplored}
We see three reasons. (\textit{i}) Cooperative MARL adopted the Dec-POMDP~\cite{oliehoek2016concise}, in which execution is decentralized by definition, and CTDE arose as a way to rehearse decentralized plans with centralized information~\cite{oliehoek2008optimal}. This was the right model when deployment-time communication was unavailable or prohibitively costly, but the absence of team-side information gradually hardened from an explicit assumption into a default, and progress concentrated on critics and value factorization. (\textit{ii}) Standard benchmarks~\cite{samvelyan2019starcraftmultiagentchallenge, papoudakis2021benchmarking} consider fixed objectives and short episodes, leaving little room for drift or stale roles, and report final return rather than recovery after change. (\textit{iii}) A coordinator adds overhead, latency, and a new failure mode. These costs are why CTME is ``\textit{mixed}", keeping local competence when coordination is delayed or unavailable. We note that application-driven work already builds central coordinators whenever the deployment permits~\cite{eisenach2024neural}.

\section{Existing Frameworks for Cooperative MARL}
\label{sec: Rel}
We review the cooperative MARL paradigms through the lens that motivates CTME. Our claim is that each existing paradigm addresses one slice of the problem, but none provides the test-time realignment mechanism the recoverability gap requires.
 
\textbf{CTDE.} The dominant paradigm~\cite{huh2023multi} uses global information during training for credit assignment, with value decomposition methods (\textit{e.g.}, VDN~\cite{sunehag2017value} and QMIX~\cite{rashid2020monotonic}) factorizing joint value functions, and centralized critic methods (\textit{e.g.}, MADDPG~\cite{lowe2017multi} and MAPPO~\cite{yu2022surprising}) providing global value estimates. These approaches stabilize learning but assume that coordination encoded during training will continue to hold throughout execution. They can neither detect when this assumption breaks nor realign the team when it does, a structural gap that better mixing networks or critics cannot close.
 
\textbf{CTCE.} At the other extreme, a fully centralized policy directly selects joint actions from the global state. CTCE guarantees coordination by construction but is exponential in team size \cite{shoham2008multiagent} and can have \textit{difficulty} accounting for the local observations and obstacles each agent faces \cite{foerster2018counterfactual, sunehag2017value, oliehoek2016concise}, eliminating the recoverability problem by removing decentralization entirely.
 
\textbf{Communication-based MARL.}
Several works learn to communicate at execution~\cite{foerster2016learning, sukhbaatar2016learning, hoshen2017vain, peng2017multiagent, jiang2018learning, singh2018learning, das2019tarmac, kim2019learning, lee2025unifying} through pairwise, broadcast, all-to-all, or server-mediated channels, also learning when to communicate~\cite{niu2021multi}, and networked or mean-field methods exploit neighborhood structure for scalability~\cite{yang2018mean}. The difference from CTME lies not in topology but in content: these channels carry agent-generated local information from which each agent reconstructs what it needs, rather than the output of a coordinator that maps team-level information to coupled high-level assignments. The two branches are complementary. Specifically, local communication can support fast neighborhood coordination and feed telemetry to a coordinator, and a communication method whose central module issues globally informed assignments is an instance of CTME.

\textbf{Hierarchical and goal-conditioned RL.} Hierarchical and goal-conditioned policies introduce higher-level abstractions such as skills, options, or goals~\cite{eysenbach2018diversity,sutton1999between, zhang2021learninginvariantrepresentationsreinforcement} to improve generalization and long-horizon performance. These methods focus on how an individual agent represents and executes sub-tasks, but typically assume that the goal or role is given. They do not address who decides which agent receives which goal, or when assignments must be revised as conditions change. CTME complements these approaches by treating goal or skill assignment as the coordinator's explicit responsibility, and by re-evaluating it as the global state evolves.

\input{table/comparison}

\textbf{The information-structure view.} Across these paradigms, the difference that matters for our position is not the choice of mixing network or critic, but the source and frequency of test-time coordination and the resulting ability to recover from drift. Table~\ref{tab:paradigm} summarizes this view. CTDE and GCRL share an absence of any test-time signal channel. Comm-MARL and CTME both add such a channel, but differ in whether the signal is emergent and P2P or explicit and globally informed. CTCE collapses the channel by absorbing all decision-making into a single controller. CTME occupies the open middle in which a coordinator broadcasts high-level signals while agents retain local autonomy.

\section{Centralized Training and Mixed Execution (CTME)}
\label{sec:ctme}

\input{figure/overview}

\textbf{CTME is a practical problem setup for test-time information structure rather than a learning algorithm.} It is parameterized by the following axes: the signal source that observes team-level state at execution, the signal abstraction (\textit{e.g.}, primitive action, sub-goal, skill, option, or latent intent), the coordination frequency (\textit{e.g.}, initial only, periodic, event-triggered, or every step), and the coordinator authority. We define CTME as a Dec-POMDP augmented with a coordinator, examine the two design problems within this space in Section~\ref{subsec:two-stage}, and show how the coordination period $k$ recovers CTDE as a limit case while keeping CTCE structurally separate in Section~\ref{subsec:freq}.

\textbf{Problem setup.}\footnote{Throughout, superscripts index agents, subscripts index timestep, and bold symbols denote joint quantities.} A team of cooperative agents must learn to solve a range of related tasks. Training is fully centralized as in CTDE, while execution is neither fully centralized nor fully decentralized. A central coordinator $C$ operates alongside the agents, observes the global state $s$, and emits signals $(z^1, \ldots, z^N)$. Each agent $i$ acts according to $\pi_i(a^i \mid o^i, z^i)$. The coordinator does not produce primitive actions but sub-goals, skills, or latent intents that each agent realizes locally. CTME does not require omniscience: the global state $s$ may be a delayed or compressed aggregate of local telemetry. What matters is the existence of a team-level information channel that no single agent can synthesize from its local observation alone.
 
We formalize CTME by combining a decentralized partially observable Markov decision process (Dec-POMDP)~\cite{oliehoek2016concise} with a coordinator $C$ as $\mathcal{M} = \langle \mathcal{I}, \mathcal{S}, \{\mathcal{A}_i\}, \mathcal{T}, \mathcal{R}, \{\mathcal{O}_i\}, \Omega, \gamma, C, \mathcal{Z} \rangle.$ Here, $\mathcal{I}=\{1,\ldots,N\}$ denotes the agent set, $\mathcal{S}$ the state space, $\mathcal{A}_i$ the action space of agent $i$, $\mathcal{T}(s'\mid s, \mathbf{a})$ the transition function for the joint action $\mathbf{a}=(a^1,\ldots,a^N)$, $\mathcal{R}(s,\mathbf{a})$ the shared reward, $\mathcal{O}_i$ the observation space of agent $i$, $\Omega(o^i\mid s)$ the observation function, $\gamma \in [0,1)$ the discount factor, and $\mathcal{Z} = \prod_i \mathcal{Z}_i$ the joint signal space. Crucially, $C$ is not a standard RL agent. It does not act on the environment; it is a separate process that emits signals based on the global state $s$.

\subsection{Two design problems within CTME}
\label{subsec:two-stage}
Within this design space, any instantiation must address two design problems. One is acquiring a diverse behavioral foundation. The other is learning a coordinator that aligns it with team objectives. The canonical two-stage pipeline below makes this decomposition explicit, although joint end-to-end training of agents and a coordinator is also consistent with CTME and may be preferred when the skill structure is not known in advance.
 
\textbf{Stage I: Behavioral foundation.} Each agent learns a signal-conditioned policy $\pi_i(a^i \mid o^i, z^i)$. This policy expresses a diverse repertoire of distinguishable behaviors indexed by $z^i\in\mathcal{Z}_i$. The objective is not to solve any specific task. It is to build a behavioral foundation. Several methods can instantiate this stage via unsupervised or self-supervised objectives, \textit{e.g.}, skill discovery~\cite{eysenbach2018diversity}, successor features for task-agnostic representations~\cite{agarwal2024proto}, and goal-conditioned pretraining~\cite{chuck2025nullcounterfactualfactorinteractions}.
 
\textbf{Stage II: Coordinator alignment.} A coordinator is introduced to align competent agents toward a specific team objective. The coordinator policy $\pi_C(\mathbf{z} \mid s)$ is trained in team reward while the agent policies remain frozen. This restores stationarity from the coordinator's perspective. Agents' responses to latents are fixed. The coordinator then faces a tractable single-agent or low-dimensional joint problem in the latent space. Whether agents receive jointly sampled or shared latents is a design choice. The latter dramatically reduces the coordinator's action space but risks losing expressivity. 

\subsection{How frequently does the coordinator activate?}
\label{subsec:freq}
\textbf{Periodic coordination.} Decentralized agents act every timestep by sampling $a^i_t \sim \pi_i(\cdot \mid h^i_t, z^i_t)$, where $h^i_t$ is agent $i$'s local history and $z^i_t$ is the most recent signal it received. The coordinator activates every $k$ steps: $\mathbf{z}_t \sim \pi_C(\cdot \mid s_t)$ if $t \equiv 0 \pmod{k}$, and $\mathbf{z}_t = \mathbf{z}_{t-1}$ otherwise. Between activations, the agents continue to act on the previous signal. The period $k$ controls how often the team is realigned. Small $k$ means more frequent realignment and higher coordination cost. Large $k$ means agents act more autonomously between updates.
 
\textbf{Coordination frequency of CTME.} CTME forms a continuous spectrum indexed by the coordination period $k$. As $k\to\infty$, the coordinator fires only once, at deployment, which amounts to CTDE with an initial role assignment; removing even this signal, or making it uninformative, recovers CTDE exactly. CTCE, by contrast, is not reached by varying $k$ alone. At $k=1$, the coordinator fires every step, but CTME continues to factor action selection across agents through $\pi_i(a^i \mid o^i, z^i)$. CTCE instead learns a single joint policy $\pi(\mathbf{a}\mid s)$ over the joint action space $\prod_i \mathcal{A}_i$. The two coincide only when each $z^i$ encodes a primitive action, which is a degenerate choice of signal space. Therefore, CTME places CTDE strictly inside its design space while keeping CTCE structurally separate. The open middle of CTME at finite $k$ is the region this paper argues should receive direct attention.
 
\textbf{Beyond periodic coordination.} Asynchronous variants in which the coordinator activates on event triggers, either coordinator-side $\beta_C(s_t, \mathbf{z}_t)\in\{0, 1\}$ or agent-side $\beta_i(h^i_t, z^i_t)\in\{0, 1\}$, inspired by the options framework~\cite{sutton1999between}, are a natural extension, but complicate training.
 
\section{Minimal Examples of Recoverability Failure}
\label{sec:analysis}

We construct a didactic bottleneck game and analyze it in two regimes that isolate distinct components of the recoverability gap. The static regime shows that anonymous decentralized execution lacks the information channel needed to select among optimal joint outcomes; we do not claim that CTDE fails in general. The sequential regime shows that even when symmetry can be broken offline, hidden global non-stationarity makes the required reassignment a function of state that no agent observes locally. Together, these regimes isolate the two failure modes test-time coordination is meant to address: static symmetry-breaking and state-contingent reassignment.
 
\subsection{Didactic example: $N$-agent symmetric bottleneck game}
 
$N$ homogeneous agents must each choose one of $K=N$ resource lanes, $a^i\in\{1,\ldots,K\}$. Each lane has a capacity of $1$. Let $n_\ell(\mathbf{a})$ denote the number of agents on lane $\ell$ under the joint action $\mathbf{a}$. The team reward is shared across all agents and is defined as follows:
\begin{equation}
R(\mathbf{a}) = \sum_{\ell=1}^{K} \big[ \mathbbm{1}\{n_\ell(\mathbf{a}) = 1\} - c\, n_\ell(\mathbf{a})\, \mathbbm{1}\{n_\ell(\mathbf{a}) \geq 2\} \big], ~ \text{where}~ u_i(\mathbf{a}) =
\begin{cases}
+1 & \text{if } n_{a^i}(\mathbf{a}) = 1, \\
-c & \text{if } n_{a^i}(\mathbf{a}) \geq 2.
\end{cases}
\end{equation}
Equivalently, $R(\mathbf{a}) = \sum_{i} u_i(\mathbf{a})$. A successful lane adds $+1$, and a colliding lane subtracts $c$ for each agent on it. This yields a single team reward shared by every agent at every step, consistent with the cooperative MARL setting. The optimum is $R^* = N$, attained by any permutation assignment.
 
\subsection{Anonymous CTDE confines welfare to the symmetric Nash equilibrium}
\input{figure/scaling}
We focus on the anonymous, exchangeable setting in which homogeneous agents cannot distinguish themselves a priori. Under decentralized execution in this setting, identical observations yield identical policies, so the only symmetric Nash equilibrium is uniform mixed play. A natural objection is that agent IDs or fixed pre-assigned roles solve the symmetry-breaking problem. They do solve the static version: if the optimal assignment is fixed, agents can learn a convention. However, this does not address the recoverability problem we target. Once the optimal mapping becomes a function of a hidden global state that no single agent observes, fixed roles become stale and the team cannot reassign without a globally informed signal. The toy example below isolates the static failure cleanly, and the sequential extension in Section~\ref{subsec:nonstat} shows the dynamic failure that fixed roles cannot fix.
 
\begin{proposition}[NE Welfare]
\label{prop:ne_welfare}
Under the symmetric NE, the expected social welfare is
\begin{equation}
W_\text{NE}(N, c) = N \cdot \left[\left(1 - \tfrac{1}{N}\right)^{N-1} - c \cdot \left(1 - \left(1 - \tfrac{1}{N}\right)^{N-1}\right)\right].
\label{eq:wne}
\end{equation}
As $N \to \infty$, the per-agent welfare $W_\text{NE}(N, c)/N$ converges to $1/e - c(1 - 1/e)$, which is negative for $c > 1/(e-1) \approx 0.582$.
\end{proposition}
Figure~\ref{fig:scaling} (\textbf{a}) confirms that the collision probability under the symmetric NE, $1-(1-\nicefrac{1}{N})^{N-1}$, quickly approaches $1 - 1/e$ as the team grows. This is not an artifact of poor optimization but a structural property of the equilibrium concept under anonymity, in which a constant fraction of agents collide in expectation.

\subsection{How does CTME unlock the correlated equilibrium?}

\input{figure/ne-ce-cce}

A coordinator drawing from the uniform distribution over the permutations of $\{1, \ldots, N\}$ and sending each agent its assigned lane achieves zero collisions and $W_\text{CE} = N$. The structure of this signaling policy mirrors a CE: a joint action is drawn from a common distribution, and each agent privately observes only its own component. In our setting, however,coordinator selects its policy for team welfare rather than to satisfy strategic incentive constraints.
 
Figure~\ref{fig:scaling} (\textbf{b}) contrasts the two equilibrium concepts across team sizes for three collision-cost regimes. The structural failure of anonymous CTDE is most striking for $c \geq 1$. The NE welfare diverges to $-\infty$ as the team grows, while CTME welfare remains optimal. This is not a marginal improvement, but the difference between a functioning and a non-functioning system.
 
\begin{proposition}[Scaling of Coordination Gain]
\label{prop:scaling}
For any fixed $c > 0$, the coordination gain $\Delta W(N) = W_\text{CE}(N) - W_\text{NE}(N, c) = \Theta(N)$ grows linearly in $N$. The per-agent gain $\Delta W / N$ converges to the positive constant $(1+c)(1 - 1/e)$.
\end{proposition}
 
Figure~\ref{fig:ne-ce-cce} \textbf{(a)} confirms that the per-agent benefit of coordination is bounded below by a positive constant for every $c$. This result refutes an intuition that ``in large teams, coordination averages out.'' As the number of lanes grows with $N$, so does the number of competitors for each lane, and these two effects nearly cancel. Each agent's collision probability stays between $1/2$ and $1-1/e \approx 63\%$ for every team size (Figure~\ref{fig:scaling} \textbf{(a)}). The coordinator eliminates these collisions, raising the per-agent payoff by $1+c$ with that probability. Therefore, the benefit depends on $c$ but only weakly on $N$.
 
Figure~\ref{fig:ne-ce-cce} \textbf{(b)} extends the analysis to asymmetric payoffs, confirming that the NE-CE gap is robust under asymmetry. The best CE and CCE achieve the same welfare in this game; with richer asymmetry, \textit{e.g.}, heterogeneous agents with conflicting preferences, CCE can strictly improve on CE.

\subsection{Coordination frequency-welfare trade-offs under non-stationarity}
\label{subsec:nonstat}
We now turn to the second failure mode, where the optimal mapping from agent to role changes over time and is not locally observable. Even teams with agent IDs and fully trained policies cannot recover from this kind of drift without a global signal, since the required reassignment depends on a state variable that no agent sees. We analyze how the coordination frequency $k$ interacts with environmental non-stationarity to shape welfare, adaptation, and robustness of CTME. To do this, we expand the toy example to a sequential setting with $N = 4$ agents, collision cost $c = 1$, and episode length $T = 200$. Each task is a latent permutation $\sigma$, and allowed lanes are shifted at every $\tau = 25$ steps. The coordinator observes $\sigma$ and, every $k$ steps, sends each agent its assigned skill $z^i = \sigma(i)$.

\input{figure/ctme-exp}

Figure~\ref{fig:ctme-exp} reveals four facets of the coordination frequency-welfare relationship. In Figure~\ref{fig:ctme-exp} \textbf{(a)}, CTME spans a continuous range: at $k \to \infty$, agents receive only an initial latent and act independently thereafter; at $k = 1$, the coordinator fires every step resembling CTCE but retaining the hierarchical coordinator-agent separation. In this illustrative setting, the tested CTME configurations outperform the CTDE baselines, suggesting that even infrequent test-time coordination can provide value when global drift is not locally observable. In Figure~\ref{fig:ctme-exp} \textbf{(b)}, for $k \leq \tau$, the coordinator reacts to every task switch; for $k > \tau$, it misses transitions, leaving agents on stale assignments. This does not prescribe a universal $k^*$; it identifies the domain's rate of change as the relevant scale for setting $k$. Figure~\ref{fig:ctme-exp} \textbf{(c)} shows that recovery time after task switches is $\Theta(k)$, not $\Theta(\tau)$. Agents do not re-learn; the coordinator issues a new signal, and the pre-trained behavioral competence executes it. Finally, Figure~\ref{fig:ctme-exp} \textbf{(d)} shows that agents continue with the last-received signal when the coordinator fails. Relative welfare loss decreases with $k$, \textit{i.e.}, systems that rely less on the coordinator lose less when it disappears. Agents retain Stage 1 behavioral competence regardless of coordinator status, a property that follows from the separation of concerns and is unavailable to monolithic CTCE.

\section{Alternative Views}
\label{sec:alter}
\textbf{View I. Hierarchical and goal-conditioned MARL already provides high-level coordination.}
One can argue that hierarchical MARL already supplies whatever high-level coordination CTDE lacks, through skill- or goal-conditioned hierarchies \citep{yang2019hierarchical,barreto2017successor,touati2021learning,eysenbach2022contrastive}, subtask-allocation architectures \citep{iqbal2022alma}, feudal or master-slave manager-worker structures \citep{ahilan2019feudal,kong2017revisiting}, centralized learned messaging \citep{gupta2025hammer}, consensus-based latent coordination \citep{feng2024hierarchical}, and safety-oriented hierarchical controllers \citep{ahmad2025hmarl}. These approaches introduce high-level abstractions such as skills, goals, or messages, and several also include some form of centralized module. Therefore, one might argue that the recoverability gap is already addressed in pieces and that CTME merely repackages existing literature. We claim that CTME's contribution is not a new architecture but an information-structure axis that organizes these methods and makes their test-time properties explicit design choices.

We agree that these methods are valuable, and many of them complement our position rather than serve as alternatives to it. Our claim is more specific. Existing hierarchical MARL methods typically emphasize one of two ingredients without combining them. The first ingredient is a rich behavioral substrate, in which agents learn reusable low-level abstractions such as latent-skill repertoires \citep{eysenbach2018diversity,laskin2022cic,park2023metra}, successor features \citep{barreto2017successor,agarwal2024proto}, or goal-conditioned policies \citep{eysenbach2022contrastive,chuck2025nullcounterfactualfactorinteractions,lee2025temporal}. These abstractions give agents a useful set of executable behaviors, but the high-level choice of which behavior to execute is typically left decentralized at deployment or learned under the CTDE assumption. As a result, the team still lacks a globally informed channel for revising those choices when objectives shift, bottlenecks emerge, or role assignments grow stale.

The second ingredient is high-level guidance. Other hierarchical methods introduce allocators, managers, master modules, or learned messengers that influence lower-level agents through goals, assignments, messages, or other abstract signals. These methods demonstrate the value of high-level coordination, but the coordination mechanism is usually tied to a specific algorithmic template, \textit{e.g.}, composite subtask allocation, pre-specified subgoal-to-reward mappings, action-level gating, or learned message passing. Consequently, the high-level module is not treated as a general test-time information channel whose signal abstraction, activation frequency, and failure behavior become part of the cooperative MARL problem definition.

CTME isolates an underexplored combination: a reusable behavioral foundation that exposes a compact high-level action space, and a globally informed coordinator that intermittently realigns the team at test time. This separation assigns behavioral competence to the low-level policies and state-contingent realignment to the coordinator, and it shifts evaluation from static learning performance to test-time recoverability metrics such as coordination frequency $k$, recovery time after drift, stale-assignment failure, and robustness to coordinator dropout. In short, hierarchical methods ask what abstractions should be learned and how agents execute them, while CTME asks who selects those abstractions at deployment, what global information informs that selection, and how often it should be revised. In stationary domains or where local observations suffice to infer the relevant team state, decentralized hierarchical policies may be enough; in long-horizon cooperative systems with shifting objectives and emerging bottlenecks, removing the coordinator forces agents to recover from global drift using only local information.

We provide other valid alternative views and corresponding arguments in Appendix \ref{app:alternative}.

\section{Call to Actions}
\label{sec:actions}
\paragraph{Benchmarks.} Existing cooperative MARL benchmarks~\cite{samvelyan2019starcraftmultiagentchallenge, papoudakis2021benchmarking, huh2023multi} have advanced the field but tend to focus on short horizons, modest team sizes, and discrete action spaces. They leave open how cooperative MARL behaves under the conditions described in real-world applications, namely long horizons where decentralized agents drift out of alignment, large-scale team where uncoordinated local optima compound into global congestion, lifelong operation where frozen behavioral foundations carry across tasks while only the coordinator adapts, and mixed-motive cooperation with fairness or safety constraints where a coordinator can act as a mediator. We encourage benchmarks that parameterize the rate of objective change, the degree of inter-agent coupling, the duration of operation, and the presence of safety or fairness invariants. Beyond asymptotic return, evaluation should report recoverability-centered metrics such as recovery time after objective shift, stale-assignment regret, coordination cost versus welfare, robustness under coordinator dropout, and performance under delayed or compressed global telemetry. Single-agent representation learning has produced a rich toolkit~\cite{eysenbach2018diversity, barreto2017successor, touati2021learning, eysenbach2022contrastive} that offers natural Stage-I behavioral foundations, while Stage-II coordinator design remains one of the most open empirical questions raised by CTME.

\paragraph{Theory and algorithms.} Operationalizing CTME raises questions such as partial observability at the coordinator, long-horizon planning over signal trajectories, asynchronous coordination, and the structure of the coordination gap. Insights from the options framework, feudal networks, and trajectory-level world models suggest coordinators may benefit from reasoning over multi-step plans rather than single-step assignments. A learned trigger function $\beta$ must also determine when coordination is needed, and analogs such as event-driven attention~\cite{das2019tarmac}, change-point detection~\cite{aminikhanghahi2017survey}, and interrupt mechanisms in options frameworks~\cite{sutton1999between} suggest viewing $\beta$ as a learned anomaly detector that fires when the team's trajectory departs from the coordinator's expected distribution. On the theoretical side, frameworks such as price of anarchy, potential games, and Stackelberg game theory~\cite{li2017review} offer natural lenses, with the Stackelberg framing especially apt since CTME's structural asymmetry maps cleanly onto leader-follower equilibria. We hope this position serves as a starting point for benchmarks that measure recoverability and graceful degradation, theoretical tools that characterize when test-time coordination provides value, and practical coordinators that scale to long-horizon non-stationary regimes where decentralized execution alone is insufficient.

\section{Conclusion}
\label{sec:conclusion}
We have argued that test-time coordination is a missing primitive in cooperative MARL. CTDE works when agents are tightly coupled at training and weakly coupled at deployment, but it is the wrong default for long-horizon centrally monitored systems where conditions evolve and the team must realign during execution. The gap is in the information structure at deployment, not the training algorithm. CTME names the design space that closes this gap, not as a single algorithm but as a test-time information structure parameterized by signal abstraction, authority, and coordination frequency. A coordinator emits high-level signals while agents act locally, and CTDE is recovered when the signal carries no information.
 
Many cooperative MARL domains already run with monitoring and coordination infrastructure in place. We hope CTME helps cooperative MARL engage with this infrastructure rather than abstracting it away, and that benchmarks measure recovery time and robustness under coordinator failure rather than only asymptotic return. The question is no longer whether coordination can be removed at test time, but when it is safe to remove it. CTDE teaches agents to coordinate before deployment, while CTME gives teams a way to recover coordination during deployment.

\section*{Acknowledgements}
This work was supported by the National Science Foundation Graduate Research Fellowship Program under Grant No. NSF 2340651, NSF 2402650, and ARO W911NF-24-1-0193.

{
\small
\bibliographystyle{unsrt}
\bibliography{main}
}

\newpage
\appendix
\input{appendix/appendix}

\end{document}

%% file: table/comparison.tex
% \begin{table}[t]
% \centering
% \footnotesize
% \caption{\textbf{Comparison of cooperative MARL paradigms by test-time information structure.} CTDE provides no test-time coordination channel and cannot recover from drift. CTME makes the channel explicit while preserving local autonomy. Here, GCRL refers to Goal-Conditioned RL and Comm-MARL refers to Communication-based MARL}
% \resizebox{\textwidth}{!}{\ttfamily{\begin{tabular}{lccccc}
% \toprule
% \textbf{Paradigm} & \textbf{Signal source} & \textbf{Frequency} & \textbf{Signal type} & \textbf{Recoverability} & \textbf{Scalability} \\
% \midrule
% CTDE       & None                & Training only & None              & $\times$    & High, brittle \\
% GCRL       & External (fixed)    & Initial only  & Goal              & $\times$    & High, brittle \\
% Comm-MARL  & Peer agents         & Intermittent  & Learned messages  & $\triangle$ & Tunable \\
% {\color{darkUT}CTME} & {\color{darkUT}Coordinator} & {\color{darkUT}Intermittent} & {\color{darkUT}Skill/Goal/Option} & {\color{darkUT}$\bigcirc$} & {\color{darkUT}Tunable} \\
% CTCE       & Central controller  & Every step    & Joint action      & $\bigcirc$  & Poor \\
% \bottomrule
% \end{tabular}}}
% \label{tab:paradigm}
% \end{table}

\begin{table}[t]
\centering
\footnotesize
\caption{\textbf{Execution-time information structures in cooperative MARL.}
CTME targets centrally monitored deployments, where aggregate team-level telemetry is available but primitive centralized control is undesirable. Here, latent denotes any compact conditioning signal above primitive actions, including skills, options, subgoals, roles, goals, or intents. $\checkmark$ denotes structurally available, $\triangle$ denotes partial or task-dependent, and $\times$ denotes absent.}
\ttfamily
\resizebox{\textwidth}{!}{
\begin{tabular}{lcccccc}
\toprule
\textbf{Paradigm} 
& \textbf{Exec. infra.}
& \textbf{Signal source} 
& \textbf{Signal type} 
& \textbf{Global telemetry} 
& \textbf{Local autonomy} 
& \textbf{Drift recovery} \\
\midrule
CTDE 
& Local-only 
& None 
& None 
& $\times$ 
& $\checkmark$ 
& $\times$ \\

GCRL 
& Goal-specified 
& External setter 
& Goal 
& $\triangle$ 
& $\checkmark$ 
& $\times$ \\

Comm-MARL 
& P2P or Server
& Peer or Channel
& Message 
& $\triangle$ 
& $\checkmark$ 
& $\triangle$ \\

Hierarchical MARL 
& Task-dependent 
& Manager or Planner 
& Latent 
& $\triangle$ 
& $\checkmark$ 
& $\triangle$ \\

CTCE 
& Fully centralized 
& Central controller 
& Joint action 
& $\checkmark$ 
& $\times$ 
& $\checkmark$ \\

\textbf{\color{darkUT}CTME} 
& \textbf{Centrally monitored} 
& \textbf{Coordinator} 
& \textbf{Latent} 
& $\checkmark$ 
& $\checkmark$ 
& $\checkmark$ \\

\bottomrule
\end{tabular}}
\label{tab:paradigm}
\end{table}

%% file: figure/overview.tex
\begin{figure}[h]
    \centering
    \includegraphics[width=\linewidth]{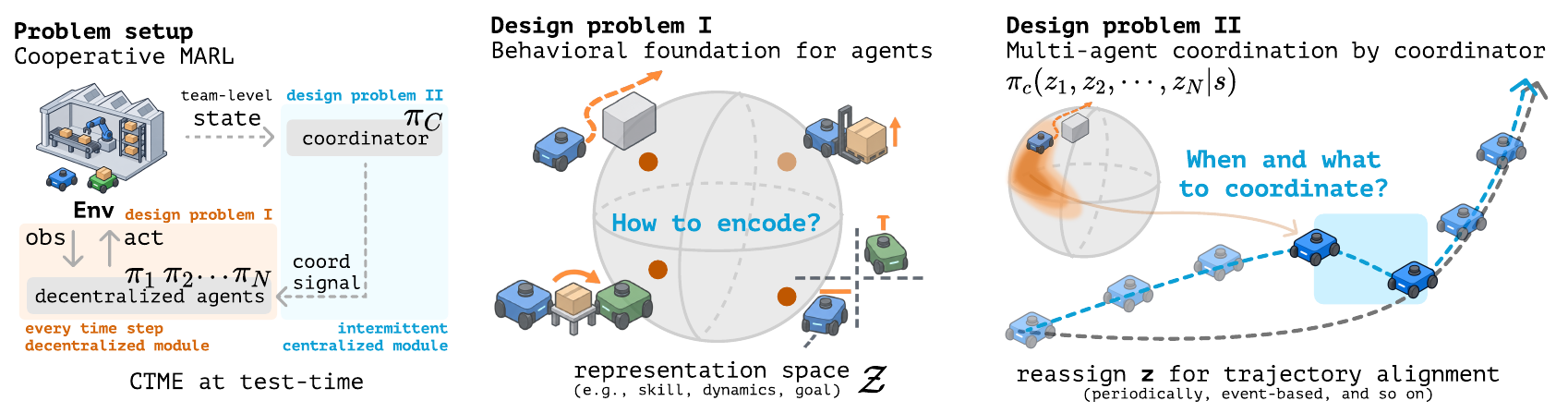}
    \vspace{-0.3cm}
    \caption{\textbf{Overview of CTME.} (\textit{Left}) Problem setup. At test time, decentralized agents act at every time step from local observations, while a coordinator intermittently maps team-level state $s$ to high-level signals $\mathbf{z}$. (\textit{Center}) Design problem I. The agent learns a behavioral foundation, a space $\mathcal{Z}$ of distinguishable behaviors such as skills, goals, or etc. (\textit{Right}) Design problem II. The coordinator selects coupled signals $\pi_C(\mathbf{z}\mid s)$ in this space. Namely, $\mathbf{z}$ is reassigned for trajectory alignment. {All fleet vehicle icons are generated by \texttt{ChatGPT-6 Astra} through multiple prompts.}}
    \label{fig:overview}
\end{figure}

%% file: figure/scaling.tex
\begin{figure}[h]
    \centering
    \includegraphics[width=\linewidth]{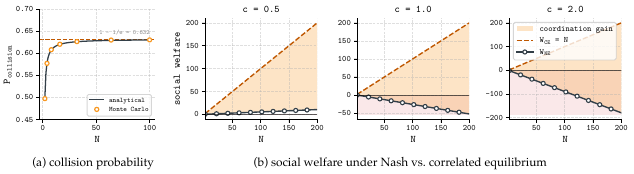}
    \vspace{-0.3cm}
    \caption{\textbf{(a)} Collision probability under the symmetric NE converges to $1 - 1/e \approx 0.632$ as $N$ increases. \textbf{(b)} Social welfare under NE (CTDE) vs. CE (CTME) for $c \in \{0.5, 1.0, 2.0\}$. The shaded coordination gain grows linearly in $N$.}
    \label{fig:scaling}
\end{figure}

%% file: figure/ne-ce-cce.tex
\begin{figure}[h]
    \centering
    \includegraphics[width=0.9\linewidth]{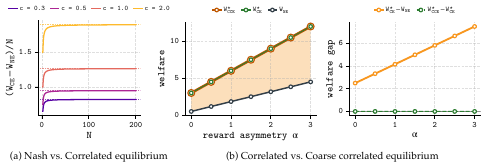}
    \vspace{-0.3cm}
    \caption{(\textbf{a}) Per-agent coordination gain $(W_\text{CE} - W_\text{NE})/N$ converges to a positive constant for each $c$. (\textbf{b}) (\textit{Left}) NE vs. CE vs. CCE welfare for the $N=3$ asymmetric game ($c=0.5$. The shaded region is the value of correlation. (\textit{Right}) The CE-NE gap grows linearly in $\alpha$, and the CCE-CE gap is zero.}
    \label{fig:ne-ce-cce}
    \vspace{-0.3cm}
\end{figure}

%% file: figure/ctme-exp.tex
\begin{figure}[h]
    \centering
    \includegraphics[width=\linewidth]{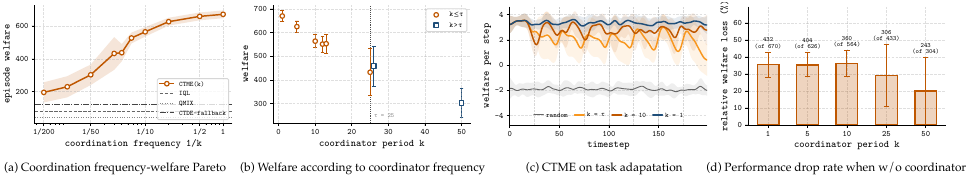}
    \vspace{-0.5cm}
    \caption{\textbf{(a)} Coordination frequency-welfare Pareto frontier. \textbf{(b)} Welfare versus coordinator period $k$. \textbf{(c)} Per-step welfare across task adaptation according to coordinator frequency $k$. \textbf{(d)} Performance drop rate of `without a coordinator' compared to `with the coordinator.' We visualize `performance without the coordinator' of `performance with the coordinator' above bar (\textit{e.g.}, $432$ of $670$).}
    \label{fig:ctme-exp}
    \vspace{-0.3cm}
\end{figure}

%% file: appendix/appendix.tex
%% appendix toc
\par\noindent\rule{\textwidth}{2pt}
\begin{table}[h]
    \vspace{-0.3cm}
    \centering
    \Large
    \begin{tabular}{c}
\textbf{Appendix}
    \end{tabular}
    \vspace{-.5cm}
\end{table}
\par\noindent\rule{\textwidth}{1pt}

\tableofcontents
\addtocontents{toc}{\protect\setcounter{tocdepth}{2}}

\newpage
\section{Extended Alternative Views}
\label{app:alternative}
\textbf{View II. CTDE with sufficient training capacity is enough.}
The most direct objection is that we may be overstating the case. Modern setups offer expressive critics, large data, and powerful architectures, and CTDE may turn out to be empirically sufficient on most tasks of interest with these tools. Our position is based on the analysis of Section~\ref{sec:analysis}. In anonymous games where the symmetric NE's collision probability approaches $1{-}1/e$, this rate reflects the equilibrium concept rather than the algorithm. Improvements in approximation quality move a learned policy closer to the equilibrium it is converging to, but do not change which equilibrium class is reachable. More precisely, more training can improve a policy within an information structure but cannot add a missing test-time channel. The information-structural gap between independent best-response policies and signal-conditioned policies is therefore unlikely to be closed by training-side innovations alone.
 
\textbf{View III. Agents can learn peer-to-peer communication.} One might argue that agents can learn to coordinate through P2P communication, without introducing a centralized coordinator. Communication-based MARL is indeed an important direction, but it does not eliminate the structural need for test-time coordination. It only changes where the coordination signal comes from.
 
P2P communication has three limitations. First, the communication protocol should be learned jointly under a shared team reward, which intensifies the credit-assignment problem \cite{foerster2016learning,foerster2018counterfactual}. A failed trajectory may result from poor behavior, poor communication, or a mismatch between the two. Second, emergent communication is often entangled with the specific agents that learned it~\cite{lowe2019pitfalls, stone2010ad}. Adding, removing, or replacing agents can break the learned protocol. Third, dense P2P communication scales poorly, \textit{i.e.}, in the worst case, each agent must reason over messages from all others \cite{sukhbaatar2016learning,das2019tarmac}.
 
CTME does not claim that P2P communication is useless. Rather, it identifies a different point in the design space. A coordinator provides a test-time coordination with access to the global context. This can reduce communication complexity, improve modularity, and separate behavioral competence from coordination competence. In this sense, CTME is not the opposite of communication-based MARL; it is a structured form of communication in which the correlation device is explicit.

\section{Scope and Deployment Considerations}
\label{app:deploy}
\textbf{What our analysis does and does not address.} CTME does not make limited-information distributed control tractable. Decentralized planning is NEXP-complete in general~\cite{bernstein2002complexity}, and CTME inherits this difficulty rather than removing it. It targets a narrower regime, where centralized training and team-level monitoring are feasible but centralized primitive-action selection is costly. Within this regime, our analysis addresses two questions: why anonymous decentralized execution cannot select among optimal joint outcomes, and how the coordination period trades welfare against coordination cost under hidden drift. The periodic model is deliberately minimal. Imperfect telemetry, asynchronous or agent-initiated triggers, unreliable or adversarial signals, and coordination at scale remain open. CTME makes them explicit design axes; it does not solve them.
 
\textbf{Deployed examples.} Industrial robot fleets show the regime most clearly. Under the VDA 5050 interface standard, robots report their status to a central fleet control that assigns orders and routes, while each robot handles localization and local motion~\cite{van2023industrial}. A central planner assigns tasks and routes, and each drive unit makes local navigation and safety decisions~\cite{wurman2008coordinating}, at a scale that recently passed one million robots~\cite{agaskar2025deepfleet}. Large-scale inventory management has a similar structure~\cite{eisenach2024neural, xie2026deepstock}. Multi-agent LLM systems, although outside our focus, share it as well. An orchestrator assigns roles and subtasks at a coarse timescale, while workers act on their own context and tool outputs~\cite{hong2024metagpt}, and inter-agent misalignment is a major failure category in such systems~\cite{li2023camel}. Drift, stale assignments, and coordination frequency arise there too.
 
\textbf{Unreliable channels and coordinator failure.} CTME does not require a reliable per-step connection. Because signals are high-level and intermittent, an agent that misses an update can keep the last valid signal or fall back to its decentralized policy. Figure~\ref{fig:ctme-exp}~\textbf{(d)} shows the extreme case of complete dropout, where welfare degrades, but agents remain operational. Robustness to noise can be built in during training by exposing agents to missing or corrupted signals and by using compact signal codes with confidence or signal-age tags. Corruption is a different matter. CTME by itself offers no robustness guarantee, and a coordinator channel is no more inherently secure than inter-agent messaging~\cite{standen2025adversarial}. Although this is not our scope and we generally assume a trusted coordinator, coordinator corruption can be an open problem as an orthogonal research direction.
 
\textbf{Relation to contextual and continual RL.} Contextual RL~\cite{jayawardana2025intersectionzoo} learns policies that generalize across context-indexed tasks, so it can instantiate the signal-conditioned local policy $\pi_i(a^i \mid o^i, z^i)$. CTME adds a deployment-time question that a fixed, externally given, or locally inferred context does not answer: how a coordinator should use team-level information to select and revise the coupled signals $(z^1, \ldots, z^N)$. Continual RL, including its multi-agent variants~\cite{nekoei2021continuous}, studies how competence is retained and expanded across a stream of tasks. CTME instead realigns the team by changing signals, without updating local policies.
 
\textbf{Adaptive triggering and Age of Information.} Periodic coordination is the simplest answer to a central design question: when should globally informed coordination be refreshed, given its cost? The Age-of-Information (AoI) literature studies this question for status updates~\cite{yates2021age}. A freshness-aware coordinator could update when the expected loss from stale telemetry or assignments exceeds the cost of realignment. Age alone is not enough, however. A signal can remain valid for a long time in a stable system, yet become wrong immediately after a drift. The Age-of-incorrect-information and value-of-information formulations~\cite{maatouk2020age} capture this by tying the updates to the correctness and the utility of the task. AoI thus informs \textit{when} to refresh, while CTME specifies \textit{what} is refreshed and how local policies use it.

\textbf{A parallel in biological motor control.} The structure of CTME is not specific to teams of physical agents. A similar organization appears inside a single body, where many local controllers must act as one team~\cite{merel2019hierarchical}. Spinal circuits and reflex loops act at fast timescales on local sensory feedback, and they can produce coordinated rhythmic stepping without input from the brain~\cite{grillner2020current}. Higher centers do not command each muscle, which would face the same combinatorial problem as CTCE~\cite{bernstein2002complexity}. Instead, they send low-dimensional signals that select or modulate behaviors, for example a brainstem command that sets the speed and gait of locomotion, or a top-down bias from the prefrontal cortex that keeps the current goal active~\cite{miller2001integrative}. These signals change at slower timescales than the local loops~\cite{murray2014hierarchy}, human control often operates intermittently rather than continuously~\cite{gawthrop2011intermittent}, and unexpected changes have been proposed to trigger a network-wide reset~\cite{bouret2005network}. These features correspond to the components of CTME: signal-conditioned local policies, a coordinator that emits high-level signals, a coordination period $k$, and an event trigger $\beta$. We use this parallel as an illustration, not as evidence. It suggests that mixed execution is a recurring solution when many coupled controllers must stay aligned under changing conditions.

\section{Proofs}
\label{app:proof}
Every agent receives the team reward $R(\mathbf{a}) = \sum_{i} u_i(\mathbf{a})$, where $u_i(\mathbf{a}) = (1+c)\,\mathbbm{1}\{n_{a^i}(\mathbf{a}) = 1\} - c$.
 
\textbf{Uniqueness of the symmetric NE.} Let all agents other than $i$ play the mixed strategy $p = (p_1, \ldots, p_N)$ over lanes, with $N \geq 2$. If agent $i$ chooses lane $\ell$, it is alone with probability $(1-p_\ell)^{N-1}$. Another agent $j$ is alone if it picks a lane $m \neq \ell$ that the remaining $N-2$ agents avoid, which happens with probability $\sum_{m \neq \ell} p_m (1-p_m)^{N-2}$. Hence, we can write as follows:
\begin{equation*}
\mathbb{E}\big[R \mid a^i = \ell\big] = (1+c)\Big[g(p_\ell) + (N-1)\textstyle\sum_{m} p_m (1-p_m)^{N-2}\Big] - cN,
\end{equation*}
where $g(x) \coloneqq (1-x)^{N-2}(1-Nx)$. Agent $i$'s best responses are thus the lanes that maximize $g(p_\ell)$. Since $g(0) = 1$ and $g(x) < 1$ for $x \in (0, 1]$, an unused lane strictly improves on any used lane. Therefore, a symmetric NE must have full support, and $g(p_\ell)$ must take a common value across all lanes. Differentiating gives $g'(x) = (N-1)(1-x)^{N-3}(Nx - 2)$, so $g$ is strictly decreasing on $[0, 2/N]$, and $g(x) \leq 0$ for $x \geq 1/N$. If some lane had $p_\ell > 2/N$, then $g(p_\ell) \leq 0$, so every other lane would need $p_m \geq 1/N$, giving $\sum_m p_m > 2/N + (N-1)/N > 1$, a contradiction. Therefore, all $p_\ell \in (0, 2/N]$, where $g$ is injective, so $p$ is uniform. Conversely, uniform play equalizes $g$ across lanes and is an NE.
 
\subsection{Proof of Proposition~\ref{prop:ne_welfare}.} 
\begin{proof}
    Under uniform play, each agent is alone with probability $q_N \coloneqq (1-1/N)^{N-1}$. By linearity of expectation, $W_\text{NE}(N, c) = \sum_i \mathbb{E}[u_i] = N\big[q_N - c(1-q_N)\big]$, which is~\eqref{eq:wne}. Since $q_N \to 1/e$, the per-agent welfare converges to $1/e - c(1-1/e)$, which is negative if and only if $c > 1/(e-1)$.
\end{proof}
 
\subsection{Proof of Proposition~\ref{prop:scaling}.} 
\begin{proof}
    A uniformly random permutation places exactly one agent on each lane, so $W_\text{CE} = N$. This distribution is a CE: an agent that deviates from its recommended lane moves onto an occupied lane and lowers the team reward. Hence $\Delta W(N) = N - W_\text{NE}(N, c) = N(1+c)(1-q_N)$. Since $q_N = (1 + \frac{1}{N-1})^{-(N-1)}$ decreases from $q_2 = 1/2$ to $1/e$, we have $1 - q_N \in [1/2,\, 1-1/e)$, so $\Delta W(N) = \Theta(N)$ and $\Delta W(N)/N \to (1+c)(1-1/e)$.
\end{proof}